\documentclass[10pt,twocolumn,letterpaper]{article}
\usepackage[T1]{fontenc}
\usepackage{wacv}
\usepackage{graphicx}
\usepackage{amsmath}
\usepackage{amssymb}
\usepackage{booktabs}
\usepackage{placeins}
\definecolor{wacvblue}{rgb}{0.21,0.49,0.74}
\usepackage[pagebackref,breaklinks,colorlinks,allcolors=wacvblue]{hyperref}

\def\wacvPaperID{28}
\def\confName{WACV}
\def\confYear{2027}

\title{5\textsuperscript{th} Workshop on Maritime Computer Vision (MaCVi): Challenge Overview}
\author{\begin{minipage}{0.96\textwidth}\centering
{\large
\mbox{Benjamin Kiefer\textsuperscript{1,2}},
\mbox{Harin Kumar Nallaguntla\textsuperscript{3}},
\mbox{Nikhil Akalwadi\textsuperscript{4}},
\mbox{Sampada Malagi\textsuperscript{4}},
\mbox{Jan Lukas Augustin\textsuperscript{5}},
\mbox{Rafia Rahim\textsuperscript{2}},
\mbox{Dominik Hildebrand\textsuperscript{2}},
\mbox{Praneel Nadgir\textsuperscript{4}},
\mbox{Abhinav Rajput\textsuperscript{6}},
\mbox{Rishii Kumar Singh\textsuperscript{6}},
\mbox{Vedant Khanna\textsuperscript{6}},
\mbox{Shubhank Gupta\textsuperscript{6}},
\mbox{Nishant Varshney\textsuperscript{6}},
\mbox{Dinesh Kumar Vishwakarma\textsuperscript{6}},
\mbox{HeeJae Kwon\textsuperscript{7}},
\mbox{Wei-Chieh Sun\textsuperscript{7}},
\mbox{Taehyun Lee\textsuperscript{7}},
\mbox{Gyungmin Ko\textsuperscript{7}},
\mbox{Yuanjun Chai\textsuperscript{7}},
\mbox{Hsiang-Wei Huang\textsuperscript{7}},
\mbox{Jenq-Neng Hwang\textsuperscript{7}},
\mbox{Tze-Hsiang Tang\textsuperscript{8}},
\mbox{Pawe{\l} Grzegrz\'{o}{\l}ka\textsuperscript{9}},
\mbox{Muhammad Waqar Mughal\textsuperscript{10}},
\mbox{Sidharth Kumar Pradhan\textsuperscript{11}},
\mbox{Sidike Paheding\textsuperscript{11}},
\mbox{Dev Thakkar\textsuperscript{12}},
\mbox{Mingi Jeong\textsuperscript{13}},
\mbox{Alberto Quattrini Li\textsuperscript{14}},
\mbox{Chaitra D Desai\textsuperscript{15}},
\mbox{Sneha Varur\textsuperscript{4}},
\mbox{Ramesh Ashok Tabib\textsuperscript{4}},
\mbox{Tusar Kanti Mishra\textsuperscript{15}},
\mbox{Sujatha C\textsuperscript{4}},
\mbox{Yi Sheng\textsuperscript{12}},
\mbox{Chen Chen\textsuperscript{16}},
\mbox{Arnold Wiliem\textsuperscript{17}},
\mbox{Matej Kristan\textsuperscript{18}},
\mbox{Uma Mudenagudi\textsuperscript{4}}\endgraf}
\vspace{4pt}
{\large
\mbox{\textsuperscript{1}LOOKOUT, Germany};
\mbox{\textsuperscript{2}University of T\"{u}bingen, Germany};
\mbox{\textsuperscript{3}University of California San Diego, USA};
\mbox{\textsuperscript{4}KLE Technological University, India};
\mbox{\textsuperscript{5}Helmut Schmidt University, Germany};
\mbox{\textsuperscript{6}Delhi Technological University, India};
\mbox{\textsuperscript{7}University of Washington, USA};
\mbox{\textsuperscript{8}Independent researcher, Taiwan};
\mbox{\textsuperscript{9}Independent researcher, Poland};
\mbox{\textsuperscript{10}Constructor University, Germany};
\mbox{\textsuperscript{11}Fairfield University, USA};
\mbox{\textsuperscript{12}University of South Florida, USA};
\mbox{\textsuperscript{13}Virginia Tech, USA};
\mbox{\textsuperscript{14}Dartmouth College, USA};
\mbox{\textsuperscript{15}Manipal Academy of Higher Education, India};
\mbox{\textsuperscript{16}University of Central Florida, USA};
\mbox{\textsuperscript{17}Shield AI, Australia};
\mbox{\textsuperscript{18}University of Ljubljana, Slovenia}\endgraf}
\end{minipage}}

\begin{document}

\AddToHookNext{shipout/foreground}{%
  \put(54,-747){\makebox(0,0)[lt]{%
    \parbox[t]{504pt}{\raggedright\fontsize{6.5}{7.5}\selectfont
      \textcopyright{} 2027 IEEE. Personal use of this material is permitted.
      Permission from IEEE must be obtained for all other uses, in any current
      or future media, including reprinting/republishing this material for
      advertising or promotional purposes, creating new collective works, for
      resale or redistribution to servers or lists, or reuse of any copyrighted
      component of this work in other works.}}}%
}

\maketitle

\begin{abstract}
The fifth MaCVi workshop at WACV 2027 reports results from three maritime vision challenges: 3D sonar diver detection (Diver3D), embedded underwater restoration and segmentation (EURS), and monocular visual odometry from boat video (LOOKOUT). We describe the datasets, scoring protocols, baselines, and final rankings, and distinguish the verification performed on each track. In Diver3D, the best entry scores 0.7602 on the Sonar--Diver Detection Score, while precise box overlap and orientation remain difficult. EURS evaluates restoration and segmentation under model-size and compute constraints on a Jetson AGX Orin. In LOOKOUT, the leading eligible entry scores 9.15\% mean drift across all 20 test clips, compared with 28.02\% for an image-free constant-velocity reference. We discuss the effects of test-label exposure, leaderboard feedback, and partial reproduction on the interpretation of the results. Challenge resources are available at \url{https://macvi.org/workshop/macvi27}.
\end{abstract}

\section{Introduction}
Maritime environments, with their dynamic lighting, reflections, waves and limited underwater visibility, require robust perception systems. Autonomous surface and underwater platforms must detect relevant objects, estimate geometry and motion, and operate under practical sensing and computation constraints. Standardized benchmarks and a collaborative research community are essential for measuring progress.

MaCVi at WACV 2027 builds on the preceding workshop series~\cite{Kiefer_2023_WACV,Kiefer_2024_WACV,Kiefer_2025_WACV,Kiefer_2026_CVPR}. The fifth edition brings together three complementary challenges: sonar-based 3D diver detection, embedded underwater image restoration and semantic segmentation, and visual odometry from maritime visible, low-light and thermal video. Together, these tasks examine object geometry, image and scene understanding, and ego-motion under sensing and deployment constraints. Table~\ref{tab:overview} summarizes their different inputs, outputs and evaluation settings.

\begin{table*}[t]
\centering
\small
\begin{tabular}{@{}p{0.12\textwidth}p{0.22\textwidth}p{0.24\textwidth}p{0.32\textwidth}@{}}
\toprule
Challenge & Input & Required output & Evaluation focus \\
\midrule
Diver3D & Sparse 3D sonar point clouds & Fully oriented 3D diver boxes and confidence scores & Exact volumetric overlap; detection, translation, scale and orientation quality \\
EURS & Degraded underwater images & Restored images and five-class semantic predictions & Joint restoration and segmentation; organizer-side embedded execution for Track 2 \\
LOOKOUT & Visible, low-light or thermal video, calibration and timestamps & Timestamped 3D camera positions & Similarity-aligned positional drift, trajectory coverage and tracking failures \\
\bottomrule
\end{tabular}
\caption{Overview of the WACV 2027 challenge suite. The tracks use different primary metrics and completion requirements; organizer baselines are reported separately from eligible participant results.}
\label{tab:overview}
\end{table*}

\section{Challenge Participation Protocol}
Teams registered and submitted through the workshop website. The track pages provide data, instructions, and evaluation tools. The following sections describe the evaluated data and metrics, distinguish development scores from held-out results, and state how each track selected and verified its final entries. Organizer baselines are reported separately from ranked participant entries.

\section{MaCVi 2027 Challenge Tracks}
\subsection{Diver3D: 3D Sonar Diver Detection}
\label{sec:diver3d}
Diver3D addresses diver detection in sparse 3D sonar point clouds. Each input contains Cartesian coordinates and intensity; a prediction consists of a box center, length, width, height, full 3D rotation and confidence. The task has one foreground class, diver. Unlike upright-box detection, it requires arbitrary rotation in $SO(3)$ rather than yaw alone. Temporal methods are permitted, while predictions are evaluated per frame.
\subsubsection{Dataset}
\begin{figure}[t]
\centering
\includegraphics[width=\linewidth]{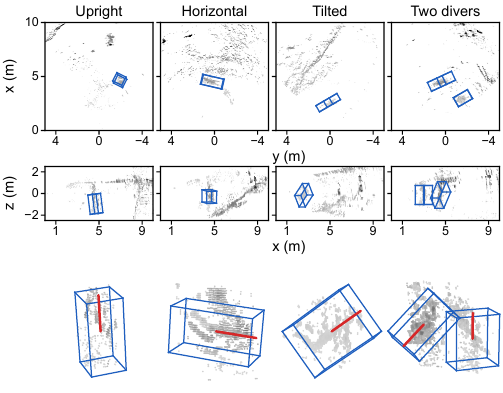}
\caption{Example Diver3D training frames (columns): an upright, a horizontal and a tilted diver, and a frame with two divers. Rows show the top view and the side view of the sonar point cloud (darker points: stronger return) with the annotated boxes, and a 3D close-up of each diver with its box and head direction (red).}
\label{fig:diver3d_samples}
\end{figure}

\paragraph{Scenes and annotations.} Diver3D is derived from uScenes~\cite{dong2026uscenes}, recorded over seven field sessions at Blue Grotto, Florida; the challenge release\footnote{\url{https://macvi.org/workshop/macvi27/challenges/diver3d}} contains only the sonar point clouds, with the full-$SO(3)$ diver annotations of SonarVoxNet~\cite{park2026sonarvoxnet}. It has 36 scenes, 37,010 frames and 34,568 diver annotations. Four divers appear in the dataset, with identities overlapping across splits; the benchmark should therefore not be described as an unseen-diver identity test. Empty frames, which comprise 27.6\% of the dataset (31.6\% of the test split), are retained as genuine negative examples. Data are released under CC BY-NC 4.0 and the tools under MIT. Figure~\ref{fig:diver3d_samples} shows example training frames with their annotations.

\paragraph{Splits.} Scene-level partitioning yields 21 training scenes with 20,861 frames, seven validation scenes with 8,037 frames, and eight test scenes with 8,112 frames. The corresponding annotation counts are 19,482, 7,509 and 7,577. This split differs from the one in~\cite{park2026sonarvoxnet}. Training and validation annotations are public; test annotations remain private.

\paragraph{Test-label exposure.} The test annotations were exposed twice during the challenge. On 4 October 2026, before submissions opened, an organizer archive containing the test labels was accessible via its link for 44 minutes. From 5 to 9 October, a public code repository of the reference method~\cite{park2026sonarvoxnet} contained the annotations of all scenes, including the test split, together with checkpoints trained on part of the test scenes. The labels were removed from the repository history on 9 October and the repository was later made private. The organizers kept the test split rather than replace it during the challenge, so the final ranking uses the original test labels. We have no evidence that any participant used the exposed labels or checkpoints, and no entry comes close to oracle accuracy (the ground truth scores SDS 1.0, the best entry 0.760), but the exposure cannot be ruled out, and the results should be read with this caveat.

\subsubsection{Evaluation Protocol}
Boxes use metric sonar coordinates with $x$ forward, $y$ left and $z$ up. Evaluation uses exact 3D intersection-over-union (IoU) of fully oriented cuboids, not a bird's-eye-view approximation. Predictions are matched greedily by descending confidence, separately at each IoU threshold, and average precision is reported at 0.30, 0.35, 0.40 and 0.50. The primary mean average precision, $m$, averages the first three thresholds. Translation error $e_t$, scale error $e_s$, and full-rotation geodesic orientation error $e_r$ are averaged over true positives matched at IoU $\geq 0.35$; $e_r$ is minimised over a $180^\circ$ turn about the diver's head-to-toe axis. The Sonar--Diver Detection Score is
\begin{align}
q_t &= 1-\min(1,e_t/1\,\mathrm{m}), \nonumber\\
q_s &= 1-e_s, \qquad
q_r = 1-\min(1,e_r/90^\circ), \nonumber\\
\mathrm{SDS} &= (3m+q_t+q_s+q_r)/6.
\end{align}
Scale error is one minus box IoU after aligning center and orientation. Only the 100 most confident boxes per frame are scored, frames without predictions count as empty, and a submission without true positives receives $q_t=q_s=q_r=0$. Ties are broken by AP at IoU 0.50, then by orientation error. The server scores every upload with the scorer shipped in the dataset package (SHA-256 \texttt{8944d861\dots2047}).

\subsubsection{Baselines}
No organizer baseline is ranked on the challenge split. The SonarVoxNet results~\cite{park2026sonarvoxnet} use a different split, and its released checkpoints were trained on four of the eight challenge test scenes, so neither is comparable with participant entries.

\subsubsection{Submissions, Analysis and Results}
Submissions were accepted from 4 to 27 October 2026, 23:59 AoE, and every upload was scored on the full test split; there was no separate testing phase. The public leaderboard lists 107 scored submissions made before the deadline by 13 participants (eight registered teams and five accounts without a team); the only later upload repeats an earlier file and is not counted. Each team is ranked by its best submission made before the deadline. Table~\ref{tab:diver3d_ranking} gives the four highest-ranked teams.

\begin{table}[t]
\centering
\small
\setlength{\tabcolsep}{1.8pt}
\begin{tabular}{@{}clcccccc@{}}
\toprule
\# & Team & SDS & $m$ & AP@.50 & $e_t$ (m) & $e_s$ & $e_r$ ($^\circ$) \\
\midrule
1 & APC\_TDC    & \textbf{0.7602} & \textbf{0.791} & 0.428 & 0.171 & 0.302 & 30.6 \\
2 & StagAI      & 0.7459 & 0.771 & \textbf{0.466} & 0.164 & \textbf{0.294} & 34.1 \\
3 & Dev Thakkar & 0.7457 & 0.766 & 0.430 & 0.169 & 0.307 & 31.3 \\
4 & FAU CAAI    & 0.7290 & 0.717 & 0.407 & \textbf{0.163} & 0.309 & \textbf{27.4} \\
\bottomrule
\end{tabular}
\caption{Final Diver3D ranking on the test split (8 scenes, 8{,}112 frames): the four highest-ranked teams, each with its best submission before the deadline (entries 19267, 19413, 19218 and 18866). SDS ranks (higher is better); $m$: mAP\textsubscript{3D} over IoU 0.30--0.40; AP@.50: AP\textsubscript{3D} at IoU 0.50; $e_t$, $e_s$, $e_r$: translation, scale and orientation errors of true positives (lower is better).}
\label{tab:diver3d_ranking}
\end{table}

The winners are APC\_TDC (SDS 0.7602), StagAI (0.7459) and Dev Thakkar (0.7457); second and third place are separated by 0.0002 SDS. APC\_TDC has the highest mAP\textsubscript{3D}, StagAI the highest AP\textsubscript{3D}@0.50 and the lowest scale error, and FAU CAAI the lowest translation and orientation errors.

\begin{figure}[t]
\centering
\includegraphics[width=\linewidth]{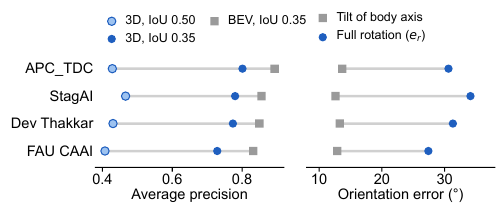}
\caption{Diver3D test results of the four highest-ranked teams. Left: average precision at a loose (0.35) and a strict (0.50) IoU threshold, in 3D and in the bird's-eye view (BEV). Right: full-rotation orientation error $e_r$ against the tilt error of the head-to-toe axis alone; the gap is rotation about the diver's long axis.}
\label{fig:diver3d_results}
\end{figure}

Localisation at a loose overlap is largely solved, while precise boxes are not (Fig.~\ref{fig:diver3d_results}). The top four reach AP\textsubscript{3D} of 0.73--0.80 at IoU 0.35, but only 0.41--0.47 at IoU 0.50. Bird's-eye-view AP at IoU 0.35 is 0.08--0.10 higher than 3D AP, so part of the remaining error is vertical. The error terms barely separate the four teams (translation error 0.163--0.171\,m, scale error 0.294--0.309), so the ranking is decided by $m$, which carries half of the SDS weight. Orientation is the weakest component: the geodesic error is 27--34$^\circ$, against 12.6--13.7$^\circ$ for the direction of the head-to-toe axis alone, so most of the rotation error is a turn about the body's long axis, about which a sparse sonar return of a diver is nearly symmetric. The four teams made 83 of the 107 submissions, and the two highest scores were reached in the last three days before the deadline.

\textbf{Top three methods.} APC\_TDC and Dev Thakkar describe their methods (see the Diver3D technical reports in the supplement); StagAI provided no method description. The two described methods are similar. Both are CenterPoint-style~\cite{yin2021centerpoint} detectors on a voxelised $12\times10\times5$\,m crop, with a bird's-eye-view backbone and a 6D rotation head~\cite{zhou2019continuity} that yields a full $SO(3)$ pose, trained with losses invariant to the same $180^\circ$ body symmetry as the metric. Both copy-paste divers in the sonar's beam space, removing the echoes a pasted diver would occlude, fuse several models with weighted box fusion~\cite{solovyev2021wbf}, refine boxes along tracks, and train on the training split only, without external data or pre-training. Their temporal steps use future frames, so both are offline methods. Post-processing matters more than ensembling: for APC\_TDC, five-frame input and track-level refinement raise a single-frame model from 0.623 to 0.752 SDS, while removing the ensemble costs 0.007. Dev Thakkar discloses that ensemble members were chosen partly with leaderboard feedback.

\paragraph{Validation.} All entries were scored by the organizer server with the released scorer, and the ranking uses these scores directly. Deadline compliance follows from the order of uploads in the server log. APC\_TDC released its code, trained weights and raw test predictions; Dev Thakkar provided a method description and offered code and weights on request. The organizers did not rerun any participant method.

\subsection{EURS: Embedded Underwater Restoration and Segmentation}
EURS studies restoration of degraded underwater images together with semantic segmentation of marine life. The five semantic classes are background, coral, fish, elasmobranch and other. The task couples visual restoration with downstream semantic utility rather than treating image appearance as a sufficient outcome.


\subsubsection{Dataset}
\begin{figure}[t]
\centering
\includegraphics[width=\linewidth]{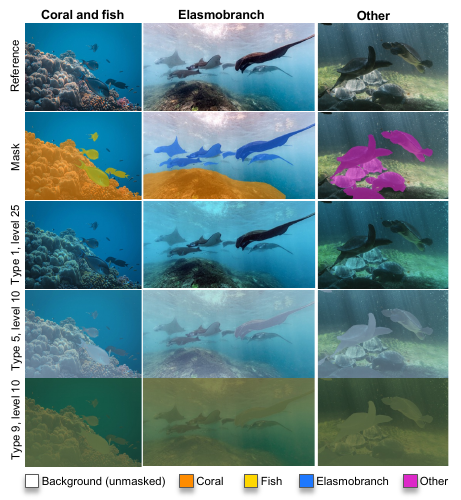}
\caption{Example EURS training scenes (columns). Rows show the clean reference, the five-class segmentation mask (overlaid; background unmasked) and three of the 200 pixel-aligned RSUIGM degradations, labelled by water type and depth level.}
\label{fig:eurs_samples}
\end{figure}

\paragraph{Scenes and annotations.} EURS builds on real underwater photographs. Each scene has one clean reference with a segmentation mask and 200 degraded versions rendered with RSUIGM~\cite{rsuigm}, which applies the optical properties of 10 water types at 20 depth levels (5 to 100, in steps of 5). Rendering changes colour and visibility but never geometry, so all versions are pixel-aligned with the reference and share its mask. References keep their native resolution (up to 48\,MP); submissions and scoring use a longest side of at most 1024\,px. A group of expert annotators labelled polygon instances of 35 species-level names, merged into the five classes (fish is the catch-all for 26 reef-fish names). Over the 30 development scenes the pooled pixel shares at scoring size are 54.2\% background, 33.7\% coral, 7.7\% fish, 1.9\% elasmobranch and 2.5\% other; elasmobranchs appear in only seven scenes. Figure~\ref{fig:eurs_samples} shows example training scenes with their masks and three of the 200 degraded versions.
\paragraph{Development release.} The 30 development scenes split by scene into 20 training scenes (4,000 images, with references and masks), 6 development scenes (1,200 degraded images, references kept on the server for Track~1 scoring) and 4 held-out scenes, which served Track~2 before the testing phase.
\paragraph{Testing-phase data.} The testing phase uses 23 new scenes, each with a clean reference image, a segmentation mask and 200 degraded versions generated in the same way as in the development release. Scene identifiers are anonymized. Track~1 is scored on 6 of these scenes (1,200 degraded images, released without references). Track~2 is scored on a private set of 21 scenes (4,200 images, all 200 versions each), which includes the six Track~1 scenes; only the Track~1 degraded images were released.
\subsubsection{Evaluation Protocol}
Track~1 takes a zip with one restored image and one single-channel mask per released image, at a longest side of at most 1024\,px. Track~2 takes a static-shape ONNX file (input $(1,3,H,W)$ in $[0,1]$, outputs \texttt{restored\_image} and \texttt{seg\_logits}) of at most 12\,GFLOPs and 1.0\,M parameters; the pipeline compiles it with TensorRT on the device, measures throughput and energy per frame, runs it on every test image and scores the outputs. Both tracks rank on the same composite
\begin{align}
Q_{\mathrm{seg}} &= \overline{\mathrm{F1}}\cdot\overline{\mathrm{IoU}}\cdot\min\!\Big(1,\tfrac{\min_k\mathrm{IoU}_k}{0.02}\Big), \nonumber\\
Q_{\mathrm{rest}} &= 0.45\,P+0.45\,\overline{\mathrm{SSIM}}+0.10\,\overline{H}, \nonumber\\
Q &= 0.60\,Q_{\mathrm{seg}}+0.30\,Q_{\mathrm{rest}}+0.10\,Q_{\mathrm{nr}},
\end{align}
where F1 and IoU are class means from pixel counts pooled over the evaluation set (the last factor zeroes $Q_{\mathrm{seg}}$ if a present class is ignored), $P$ is the mean PSNR normalized over 15--35\,dB, $H$ is a sharpness ratio against the reference, and $Q_{\mathrm{nr}}$ averages normalized UIQM, UCIQE and CCF~\cite{panetta2016uiqm,yang2015uciqe,wang2018ccf}, kept at a small weight because such measures reward contrast stretching. Throughput, energy, GFLOPs and parameters are reported for Track~2 but not scored.

The final ranking uses reproduced testing-phase results only. With $\hat T_t$ the min-max normalized reproduced score of a team's best submission in Track~$t$,
\begin{equation}
F = 0.70\,\hat T_2 + 0.30\,\hat T_1,
\end{equation}
with ties broken by Track~2, then Track~1. Track~2 weighs more because EURS is an embedded challenge, and results not validated on an embedded device would say little.
\subsubsection{Baselines}
The organizer baseline is a small joint network: a four-stage depthwise-separable encoder shared by a restoration decoder and a segmentation decoder, with U-Net-style skip connections, exported as one ONNX graph. It has 0.052\,M parameters and 1.17\,GFLOPs at $384\times768$, about a tenth of the compute cap. It is trained on the training scenes with an L1+SSIM loss on restoration and class-weighted cross-entropy plus Dice on segmentation, using AdamW (peak learning rate $10^{-4}$, three warm-up epochs, cosine decay), 100 epochs at $256\times256$ and a fixed seed. It is a deliberately small reference: the data is scarce, and doubling the width overfit without improving the held-out score. Organizer baselines appear on the leaderboards for reference and do not compete.
\subsubsection{Submissions, Analysis and Results}
EURS ran a development phase from early September 2026 and a testing phase from 20 to 28 October 2026. In total, 779 submissions were received from 56 participants in 27 registered teams (471 in the development phase and 308 in the testing phase; Table~\ref{tab:eurs_subs}). Track~2 drew fewer development submissions than Track~1 (196 against 275), since every ONNX model is compiled and measured on the device, but more participants in the testing phase (36 against 31). Eighteen participants entered both tracks in the development phase and 26 in the testing phase. Nine teams completed the hand-in and were ranked after reproduction of their results.

\begin{table}[t]
\centering
\small
\begin{tabular}{@{}llrr@{}}
\toprule
Phase & Track & Submissions & Participants \\
\midrule
Development & 1 & 275 & 28 \\
Development & 2 & 196 & 26 \\
Testing & 1 & 182 & 31 \\
Testing & 2 & 126 & 36 \\
\midrule
Total & & 779 & 56 \\
\bottomrule
\end{tabular}
\caption{EURS submissions received per phase and track. Participants are accounts with at least one submission; the total counts each account once. Organizer accounts are excluded.}
\label{tab:eurs_subs}
\end{table}

Table~\ref{tab:eurs_ranking} gives the final ranking. The winners are Artificial Intelligence and Machine Learning Lab (AIML Lab, $F=0.9625$), UW-IPL (0.8053) and APC\_TDC (0.7778). AIML Lab has the highest Track~2 composite (0.4300) and UW-IPL the highest on Track~1 (0.7537). The Track~1 composites of the nine teams lie between 0.637 and 0.754, against 0.184 for the organizer baseline, and the Track~2 composites between 0.327 and 0.430.

\begin{table}[t]
\centering
\small
\setlength{\tabcolsep}{2.6pt}
\begin{tabular}{@{}clccccc@{}}
\toprule
\# & Team & $T_1$ & $T_2$ & $F$ & GFLOPs & fps \\
\midrule
1 & AIML Lab & 0.7391 & 0.4300 & 0.9625 & 11.2 & 32.2 \\
2 & UW-IPL & 0.7537 & 0.4012 & 0.8053 & 11.4 & 9.5 \\
3 & APC\_TDC & 0.6905 & 0.4212 & 0.7778 & 11.3 & 8.4 \\
4 & StagAI & 0.6371 & 0.4192 & 0.6270 & 10.8 & 24.9 \\
5 & GitRemoveAll & 0.6511 & 0.4095 & 0.5973 & 9.9 & 34.0 \\
6 & SAT & 0.6413 & 0.4070 & 0.5552 & 9.7 & 33.3 \\
7 & deepdrivepl & 0.6547 & 0.3972 & 0.5233 & 6.9 & 50.8 \\
8 & GIST AILAB & 0.6656 & 0.3831 & 0.4555 & 4.2 & 86.4 \\
9 & PicusLabMarine & 0.7314 & 0.3266 & 0.2425 & 11.5 & 21.9 \\
\bottomrule
\end{tabular}
\caption{Final EURS ranking. AIML Lab: Artificial Intelligence and Machine Learning Lab. $T_1$: reproduced Track~1 composite; $T_2$: Track~2 composite of the verified ONNX graph; $F=0.7\,\hat T_2+0.3\,\hat T_1$ with min--max normalized composites. GFLOPs and frames per second: Track~2 graph on the Jetson AGX Orin.}
\label{tab:eurs_ranking}
\end{table}

The tracks rank the teams very differently (Spearman correlation between the Track~1 and Track~2 composites: $-0.08$): UW-IPL, best on Track~1, places sixth on Track~2, and PicusLabMarine, third on Track~1, is last on Track~2; only Artificial Intelligence and Machine Learning Lab is in the top two of both. With min-max normalization and a 70\% weight on Track~2, small raw differences matter: the six highest Track~2 composites lie within 0.029.

Segmentation quality falls under the Track~2 limits of 1\,M parameters and 12\,GFLOPs, from $Q_{\mathrm{seg}}$ of 0.54--0.70 on Track~1 to 0.14--0.30 on Track~2 (on a larger test set, 21 scenes against 6). Jetson throughput ranges from 8 to 86\,fps and is only weakly related to accuracy (Spearman correlation with the Track~2 composite $-0.32$): the fastest graph (4.2\,GFLOPs) trails the best entry by 0.047, while the two slowest (below 10\,fps) rank second and third overall.

The approaches are similar. All nine teams use the water type and depth level in the file names in Track~1 and fit a per-degradation colour model on the training scenes; eight use a frozen DINOv3 ViT-L/16 for Track~1 segmentation, seven distil a DINOv3-based teacher into a lightweight Track~2 student, and eight add external underwater datasets. The main difference at the top of Track~1 is SAM~3: the three highest Track~1 composites belong to the only three teams using it (UW-IPL, Artificial Intelligence and Machine Learning Lab, PicusLabMarine), while the other six score 0.637--0.691. As the teams also differ in ensemble size and post-processing, this is an association, not a controlled comparison.

\begin{figure}[t]
\centering
\includegraphics[width=\linewidth]{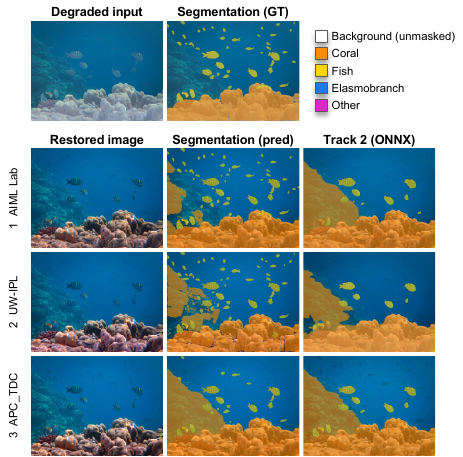}
\caption{Qualitative results on one testing scene (water type 5, level 10). Top: degraded input and ground-truth segmentation. Below, for the three highest-ranked teams: the Track~1 restored image and predicted segmentation (overlaid on the restored image), and the Track~2 output, i.e.\ the segmentation overlaid on the restored image produced by the same ONNX graph. Track~2 outputs were generated by running the submitted graphs with ONNX Runtime. Overlay colours: coral (orange) and fish (yellow). AIML Lab: Artificial Intelligence and Machine Learning Lab.}
\label{fig:eurs_top3}
\end{figure}

\subsection{LOOKOUT: Maritime Monocular Visual Odometry}
\label{sec:lookout}

\begin{figure*}[t]
\centering
\includegraphics[width=\textwidth]{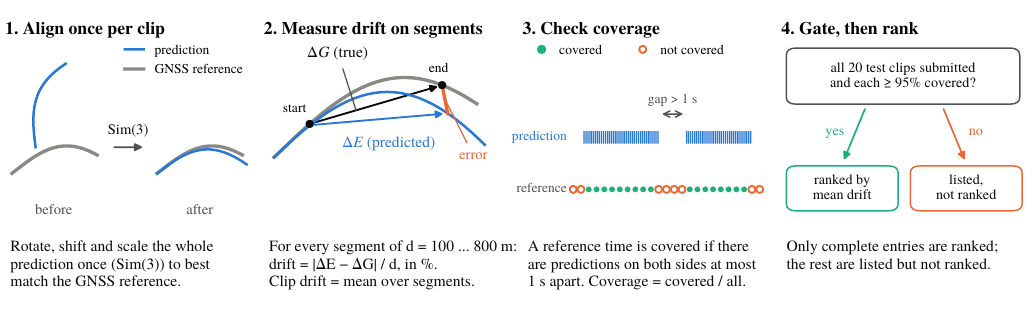}
\caption{Maritime VO evaluation in four steps (schematic, not data). Each test clip is aligned once, scored by segment drift, and checked for coverage; only entries that cover all 20 test clips are ranked.}
\label{fig:lookout_protocol}
\end{figure*}
LOOKOUT asks for the trajectory of a boat from its own monocular video, using only the video, a nominal calibration and the frame timestamps. GPS, IMU, AIS, speed, heading and attitude are withheld, since speed and heading alone would allow dead reckoning. Training uses the supplied training split; external pretraining must be disclosed, and validation or test videos may not be used for training, including self-supervised training. The setting is hard for visual odometry: open water fills much of each frame and offers little stable texture, and the horizon and distant shore give features with little parallax.

\subsubsection{Dataset}
The benchmark has 112 five-minute clips (9.3\,h, 0.83 million frames) from LOOKOUT camera systems on boats, each with a reference trajectory from the boat's GNSS. The development release\footnote{\url{https://macvi.org/workshop/macvi27/challenges/maritime-visual-odometry}} holds 68 training and 24 validation clips with references, and the test package holds 20 clips without them. Splits are formed by recording boat. The development paths cover 200.9\,km; a clip covers 0.69--7.17\,km at mean speeds of 2.3--24.0\,m/s (median 5.7\,m/s), and 54 of 92 clips have a chord at least 90\% of their path length. Six camera modes cover visible (wide, zoom, flir-VIS), low-light near-infrared (sionyx, infrared) and thermal (flir-IR) video, all at $1920\times1080$ and 14.3--30.0\,fps. Of the 112 clips, 98 are visible, 8 low-light (all in training) and 6 thermal (4 training, 1 validation, 1 test). Validation has no flir-VIS clip, while 11 of the 20 test clips are flir-VIS. Calibration is a distortion-free pinhole model from each mode's nominal field of view, with the principal point at the image centre. Curation removed clips that switch camera mode, have heavy video artifacts or people close to the camera, or have more than 3\% non-increasing timestamps. Data are released under CC BY-NC 4.0 and the starter code under MIT.


\subsubsection{Evaluation Protocol}
A prediction gives 3D positions for any increasing subset of decoded frames, in any frame and scale; orientation is not scored. References are cleaned 1\,Hz GNSS positions in a local east-north plane in metres. The evaluator interpolates the prediction onto the reference times and fits one similarity transform per clip~\cite{umeyama1991}, because monocular video observes neither metric scale nor absolute heading. Drift is measured over reference-path segments of length $d\in\{100,200,\dots,800\}$\,m, as in KITTI~\cite{geiger2012kitti}, starting every tenth reference sample (Fig.~\ref{fig:lookout_protocol}, step 2). With aligned estimate $E$ and reference $G$, a segment from $i$ to $j$ has error
\begin{equation}
e_{i,d} = 100\,\big\|(E_j-E_i)-(G_j-G_i)\big\|\,/\,d ,
\end{equation}
where $j$ is the first reference index for which cumulative path distance from $i$ reaches $d$; eligible segments span at least three reference-index intervals. Clip drift is the mean of $e_{i,d}$, with no per-segment realignment, and compares positions only. Absolute trajectory error~\cite{sturm2012tum} is reported but not ranked.

Every expected clip receives a status, and a tracking failure counts as a missing prediction. The server's gap-aware coverage gate counts a reference time as covered if a predicted sample lies at it or between two samples at most 1\,s apart (Fig.~\ref{fig:lookout_protocol}, step 3). An entry ranks only if all 20 test clips are scored and each has at least 95\% gap-aware coverage. Drift is still computed over the full overlapping window, so gaps do not remove difficult stretches from the score. The local development evaluator reports the overlap fraction as a diagnostic only. Teams nominate one eligible entry, ranked by mean drift over the 20 clips, with one evaluated upload per team per day until 27 October 2026, 23:59 AoE.

\subsubsection{Baselines}
The DPVO starter~\cite{teed2023dpvo} runs the released checkpoint (DPVO 859bbbf, SHA-256 \texttt{30d02dc2\dots0ff12}) with default settings (96 patches per frame, no loop closure) on frames cropped to the active image, sampled at about 8\,Hz and halved in resolution. A 24-clip validation run takes about 47 minutes on one RTX A5000 and writes a manifest with code revision, checkpoint hash, settings and software versions. Seeds are fixed, yet GPU execution is not deterministic: identical runs differ in drift and in which clips fail. The image-free constant-velocity reference places the camera at $(t-t_0,0,0)$ for every frame time $t$. The similarity fit then supplies heading and mean speed, so the reference is nearly exact on steady straight transits (0.5\% drift on the best validation clip) and degrades with turns and speed changes. Run manifests and summaries for these baselines are available at \url{https://drive.google.com/drive/folders/1xPVomYGVH43zogKqRx3zp3wEpZP4m-tD}.

\subsubsection{Results and Analysis}
The constant-velocity reference beats the DPVO starter. On the test leaderboard it scores 28.02\% mean drift against 58.01\% for DPVO + constant-velocity fallback; both are non-competing rows uploaded through the normal path (Table~\ref{tab:lookout_results}). The fallback entry uses one validation-setting DPVO run on 19 clips and constant velocity on clip\_096, where DPVO returned non-finite poses after about 28\,s in four of four attempts.

\begin{figure}[t]
\centering
\includegraphics[width=\linewidth]{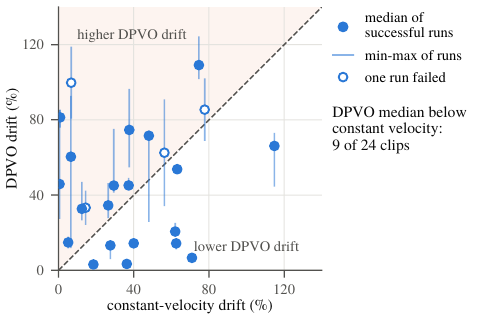}
\caption{Validation drift per clip: DPVO with identical settings in three attempted runs against the image-free constant-velocity reference. Points above the dashed line have higher DPVO drift. Medians use successful outputs; open markers indicate an incomplete set of runs.}
\label{fig:lookout_val}
\end{figure}

\begin{table}[t]
\centering
\small
\setlength{\tabcolsep}{2.2pt}
\begin{tabular}{@{}lccccc@{}}
\toprule
Method & Split & Scored & Elig. & Mean & Median \\
\midrule
Const.\ velocity & val & 24/24 & -- & 38.76 & 36.81 \\
DPVO run 1 & val & 24/24 & -- & 46.19 & 41.88 \\
\midrule
DPVO run 2 & val & 21/24 & -- & 39.68$^\dagger$ & 26.18$^\dagger$ \\
DPVO run 3 & val & 23/24 & -- & 45.36$^\dagger$ & 45.14$^\dagger$ \\
\midrule
Const.\ velocity$^\ast$ & test & 20/20 & ref. & 28.02 & -- \\
DPVO + CV$^{\ast\ddagger}$ & test & 20/20 & ref. & 58.01 & -- \\
\midrule
1.\ cyberkompaspl & test & 20/20 & yes & \textbf{9.15} & -- \\
2.\ Intelligent Systems & test & 20/20 & yes & 9.58 & -- \\
3.\ UW-IPL & test & 20/20 & yes & 10.83 & -- \\
\bottomrule
\end{tabular}
\caption{Drift in \% (lower is better). Test mean drift is the ranking metric. $^\dagger$Incomplete run: scored clips only, not comparable with 24/24 rows. $^\ast$Organizer reference, not ranked. $^\ddagger$CV = constant velocity: constant velocity on one clip where DPVO diverged. Team rows are the final entries 19385, 19264 and 19406 by leaderboard rank; every clip passes the 95\% coverage gate. Per-clip test drift is not published, so test medians are omitted.}
\label{tab:lookout_results}
\end{table}

On the full 24-clip validation set, the one complete DPVO run averages 46.19\% against 38.76\% for constant velocity, and the per-clip median over successful outputs from three DPVO runs with identical settings is below the constant-velocity drift on only 9 of 24 clips (Fig.~\ref{fig:lookout_val}). Per-clip drift moves by up to 80 percentage points between runs. The other two runs failed on 3 and 1 clips, so their means cover scored clips only. The run with three failures has the lowest such mean (39.68\%) because the clips it dropped drift heavily in the other runs, which is the scored-subset bias the complete-entry rule prevents. These results compare globally scale-aligned planar positions, so they say nothing about recovered metric scale or full six-degree-of-freedom pose. They describe this DPVO starter on a dataset rich in straight transits, with nominal calibration and no flir-VIS clips in validation.

The final leaderboard has 13 competing entries. The best three cover all 20 test clips with 9.15\% (cyberkompaspl), 9.58\% (Intelligent Systems) and 10.83\% (UW-IPL) mean drift, under 40\% of the constant-velocity reference (Table~\ref{tab:lookout_results}). The organizers rebuilt each final file on a CPU from the team's released intermediates (pose-model rotations, water-flow speeds, learned-speed predictions) to within $10^{-10}$\,m per position; the GPU stages producing these intermediates were not rerun. The LOOKOUT supplement gives each method, code and checkpoint links, and the team's disclosure of leaderboard-guided settings.

All three split motion into heading and relative speed and integrate it in the plane; the per-clip similarity alignment absorbs the unknown speed scale and initial heading. Heading comes from frozen multi-view geometry models (VGGT-1B~\cite{wang2025vggt}, VGGT-Omega~\cite{wang2026vggtomega}, Pi3X~\cite{wang2025pi3}) on short windows, once combined with bundle adjustment of far-scene tracks; the fourth-ranked entry (UNIZG FER LABUST / MARBLE, 11.02\%) also builds on VGGT-Omega and DPVO. 
Speed comes from water-surface optical flow with learned correctors trained only on the training split, and all three fall back to constant speed on the thermal test clip, where water flow carries little signal. All three disclose settings chosen with test-leaderboard feedback, two of them clip-specific; with no flir-VIS clip in validation and 11 in test, these scores include some test-set-dependent selection and may overstate performance on unseen data.

\section{Reproducibility and Results Reporting}
To make results reproducible, record evaluation versions, code and checkpoint identifiers, preprocessing, seeds, precision, and hardware. For embedded entries, also report the power mode, software stack, timing protocol, and resource limits. For trajectory methods, report completion alongside drift and identify the clips that enter each aggregate. Apply the same evaluation and retry policy across teams.

\section{Discussion and Conclusion}
Diver3D methods detect divers at loose overlap thresholds more reliably than they recover precise 3D boxes and orientation. The two top-three teams that described their methods used future frames. Their results therefore leave online performance open. EURS throughput varied widely on the Jetson and was only weakly related to the Track~2 composite, making both accuracy and speed important to report. LOOKOUT's top three covered all 20 clips and substantially improved on constant velocity under scale-aligned planar scoring. All three disclosed settings informed by test-leaderboard feedback.

Verification also differed by track. Diver3D used server-side scoring without rerunning the methods, and temporary test-label exposure limits conclusions about generalization. EURS reproduced testing-phase results and measured embedded execution on the target hardware. LOOKOUT rebuilt the top-three trajectories from released intermediates without rerunning the GPU stages. Future editions should keep test artifacts private and document exactly which steps were independently reproduced.

{\small
\bibliographystyle{ieeenat_fullname}
\bibliography{egbib}
}

\end{document}